# Dialogical Reasoning Across AI Architectures: A Multi-Model Framework for Testing AI Alignment Strategies

**Gray Cox** College of the Atlantic, Bar Harbor, Maine, USA gcox@coa.edu

## Abstract

Current approaches to evaluating AI alignment proposals rely predominantly on monological methods—single evaluators applying fixed criteria to static proposals. This paper introduces a multi-AI dialogue framework that enables structured adversarial-yet-collaborative testing of alignment strategies across different AI architectures. The framework assigns distinct roles (Proposer, Responder, Monitor, Translator) to AI systems and orchestrates multi-turn dialogues designed to surface objections, explore mechanisms, and identify areas of convergence and persistent disagreement.

We demonstrate the methodology using Viral Collaborative Wisdom (VCW), a Peace Studies-inspired alignment framework that treats alignment as fundamentally about relationship rather than control. Experimental results from 72 dialogue turns across six conditions (rotating Claude, Gemini, and GPT-4o as Proposer and Responder) reveal: (1) all three major AI architectures successfully engage with complex alignment concepts from Peace Studies traditions; (2) different architectures surface complementary objections—verification concerns from Claude, scalability concerns from Gemini, bias concerns from GPT-4o; (3) dialogue deepens through designed phases with synthesis producing novel insights rather than premature convergence; and (4) terminological precision can be maintained through explicit prompt engineering.

The primary contribution is methodological: a replicable framework for stress-testing AI alignment proposals through structured multi-model dialogue. The framework enables researchers to leverage diverse AI architectures as complementary critics, providing more comprehensive evaluation than single-model assessment.

## 1. Introduction

The challenge of ensuring that advanced AI systems remain beneficial—the AI alignment problem—has generated numerous proposed frameworks, from reward modeling and Constitutional AI to debate-based oversight and cooperative AI (Russell, 2019; Bai et al., 2022; Irving et al., 2018; Dafoe et al., 2020). Yet evaluating these proposals presents its own methodological challenge: how do we systematically stress-test alignment frameworks before deploying them in high-stakes contexts?

Current evaluation approaches are predominantly monological—a single researcher or system applies fixed criteria to assess a static proposal. This approach has significant limitations. A single evaluator, whether human or AI, brings particular blind spots and reasoning patterns. Fixed evaluation criteria may miss unexpected failure modes. And static assessment cannot capture how alignment proposals perform under sustained critique and refinement.

This paper introduces an alternative: dialogical evaluation through structured multi-AI dialogue. Drawing on Peace Studies traditions that have developed sophisticated methods for conflict transformation through dialogue (Fisher & Ury, 1981; Lederach, 1997; Buber, 1923/1970), we design a framework where different AI systems play distinct roles—defending, critiquing, monitoring, and translating—across multiple turns of structured engagement.

The key insight is that different AI architectures, trained on different data with different methods, may function as complementary critics. Just as diverse human perspectives surface different concerns in deliberative processes, diverse AI architectures may reveal different failure modes and raise different objections. Multi-model dialogue thus provides more comprehensive stress-testing than single-model evaluation.

We demonstrate this methodology using Viral Collaborative Wisdom (VCW), a Peace Studies-inspired alignment framework that proposes treating AI alignment as fundamentally about relationship rather than control (Cox, 2023). VCW draws on dialogical reasoning traditions including Fisher and Ury's principled negotiation, Gandhi's satyagraha methodology, and Ostrom's commons governance research. It serves as an ideal test case precisely because it is complex, philosophically rich, and controversial—if the methodology can facilitate substantive engagement with VCW, it should work for other alignment proposals as well.

Our experiments address three research questions:

**RQ1: Cross-Architecture Engagement.** Can different AI architectures substantively engage with complex alignment frameworks, or do they produce only superficial responses?

**RQ2: Complementary Critique.** Do different AI architectures raise different objections, providing more comprehensive stress-testing than single-model evaluation?

**RQ3: Dialogue Dynamics.** Does structured multi-turn dialogue produce genuine deepening of engagement, or does it converge prematurely to consensus or degenerate into repetition?

Our results demonstrate that the methodology succeeds on all three dimensions: Claude, Gemini, and GPT-4o all engage substantively with VCW's Peace Studies foundations; different architectures surface complementary concerns; and dialogue deepens through designed phases with synthesis producing genuinely novel positions not contained in either party's initial framing.

The paper's primary contribution is methodological rather than substantive. We do not claim to have validated VCW as an alignment approach; rather, we demonstrate a replicable framework for stress-testing any alignment proposal through structured multi-model dialogue. We provide complete prompt libraries, analysis methods, and experimental protocols to enable community replication and extension.

## 2. Background and Related Work

### 2.1 AI Alignment Approaches

The AI alignment literature has developed multiple approaches to ensuring beneficial AI behavior. Reward modeling and RLHF (Reinforcement Learning from Human Feedback) attempt to learn human preferences from comparative judgments (Christiano et al., 2017). Constitutional AI extends this by having AI systems self-critique against explicit principles (Bai et al., 2022). Debate-based approaches pit AI systems against each other to surface deception (Irving et al., 2018). Cooperative AI research investigates how to design AI systems that can collaborate effectively with humans and other AI (Dafoe et al., 2020).

These approaches share an implicit assumption: alignment is fundamentally a control problem—how do we constrain AI behavior to match human preferences? This framing treats the AI system as an object to be controlled rather than a participant in an ongoing relationship being developed through dialogical methods of reasoning including negotiation, mediation, shared problem-solving, dispute resolution, and other approaches to conflict transformation. Some researchers have questioned this control-based assumption, calling for approaches that treat alignment as collaborative rather than adversarial (Gabriel, 2020; Korinek & Balwit, 2022).

### 2.2 Monological vs. Dialogical Reasoning

VCW draws on a tradition of *dialogical reasoning* (DR) developed in Peace Studies and conflict transformation research. This approach differs fundamentally from *monological reasoning* (MR)—the linear inference from fixed premises to conclusions that characterizes formal logic and most AI reasoning systems.

| Dimension | Monological Reasoning | Dialogical Reasoning |
|---|---|---|
| **Locus** | Single reasoner applies rules to premises | Multiple perspectives encounter each other |
| **Starting point** | Premises and definitions fixed at outset | Meanings negotiated through exchange |
| **Goal** | Valid inference to conclusion | Shared understanding that transforms both parties |
| **Relation to other** | Other as object of analysis | Other as subject in I-Thou relation |
| **Success** | Winning argument or | Emergence of novel insight & genuine |

| Dimension | Monological Reasoning | Dialogical Reasoning |
|---|---|---|
| **criterion** | optimal compromise | agreement |
| **Temporal structure** | Linear: premises → conclusion | Iterative: encounter → transformation → deepened encounter |

*Table 1: Key contrasts between monological and dialogical reasoning*

Crucially, DR does not aim at compromise (splitting differences) or victory (one party persuading the other). Rather, it seeks genuine encounter through which both parties may be transformed. Buber's (1923/1970) distinction between I-It relations (treating the other as object) and I-Thou relations (genuine meeting between subjects) captures this difference: monological reasoning treats alternative perspectives as data to be processed, while dialogical reasoning treats them as subjects with whom understanding must be co-constructed.

This distinction has important implications for AI alignment. If values cannot be fully specified in advance but must be discovered through ongoing relationship, then alignment requires dialogical rather than purely monological approaches. VCW operationalizes this insight through mechanisms like the Interest Excavation Algorithm, which treats stakeholder perspectives not as fixed inputs to be aggregated but as starting points for iterative deepening of mutual understanding.

## 2.3 Peace Studies Traditions in VCW

VCW integrates several Peace Studies traditions:

**Principled Negotiation** (Fisher & Ury, 1981): The distinction between *positions* (what parties say they want) and *interests* (why they want it) enables finding solutions that satisfy underlying needs even when surface positions conflict.

**Conflict Transformation** (Lederach, 1997): Rather than merely resolving conflicts, transformation seeks to change the relationships and systems that produce conflict. This suggests alignment is an ongoing process, not a solved state.

**Satyagraha Methodology** (Gandhi): Committed action based on best current understanding, with acceptance of consequences as information for refinement. This provides an empirical method for value discovery without claiming access to absolute truth.

**Commons Governance** (Ostrom, 1990): Polycentric governance systems where multiple centers of authority interact, providing models for multi-stakeholder AI governance.

**I-Thou Dialogue** (Buber, 1923/1970): Genuine encounter between subjects, as opposed to instrumental I-It relations. This suggests that AI alignment requires treating AI systems as participants in relationship rather than objects to be controlled.

## 2.4 Multi-Agent AI Evaluation

Contemporary work on multi-agent debate in AI (Du et al., 2023; Liang et al., 2024) uses multi-agent interaction to improve performance on tasks with verifiable answers—essentially a monological goal pursued through multi-agent means. Our methodology differs in evaluating capacity for *dialogical reasoning itself*: the ability to engage constructively with alternative perspectives, negotiate meanings, and arrive at insights that emerge from relationship.

Irving et al.'s (2018) "AI Safety via Debate" proposes debate as a scalable oversight mechanism, where competing AI systems expose each other's deceptions. This is closer to our approach but remains adversarial rather than collaborative. Recent empirical work (Khan et al., 2024) demonstrates that debate with more persuasive models leads to more truthful answers, suggesting that structured AI-to-AI interaction can improve alignment outcomes.

Our contribution fills a gap in this literature: a methodology for testing whether AI systems can engage in *dialogical* reasoning about alignment itself—not merely debating factual claims, but exploring complex frameworks, surfacing diverse objections, and developing emergent understanding through sustained engagement.

## 2.5 Can LLMs Engage in Dialogical Reasoning?

Whether LLMs can engage in genuine dialogical reasoning—as opposed to sophisticated pattern matching—remains philosophically contested. Critics argue that transformer architectures are "stochastic parrots" (Bender et al., 2021) incapable of the intentionality that genuine dialogue requires.

Recent theoretical and empirical work complicates this picture. Cox (2024) argues that the Attention mechanism in transformer architectures enables forms of formal and final causality that go beyond purely mechanical next-token prediction—LLMs exhibit purposive behavior structured by anticipated outcomes, not merely causal chains from prior tokens. Empirical studies have demonstrated emergent analogical reasoning at human-level performance (Webb et al., 2023) and near-human accuracy on theory-of-mind tasks (Kosinski, 2024).

Rather than presupposing an answer to this philosophical question, our methodology offers empirical tools for investigating it. The experimental design enables observation of whether AI systems exhibit characteristic features of dialogical reasoning: mutual transformation, meaning negotiation, emergence of novel positions, and I-Thou dynamics. This approach treats the question of AI dialogical capacity as an empirical research program rather than a philosophical assumption.

## 3. Methodology

### 3.1 Experimental Design

Our framework assigns four distinct roles to AI systems:

**Proposer:** Presents and defends an alignment framework (in this case, VCW). Receives turn-specific prompts that evolve from initial presentation through response to critique toward synthesis.

**Responder:** Critically evaluates the proposed framework. Begins with genuine openness, develops substantive objections, and in later turns explores areas of potential convergence while maintaining intellectual honesty about persistent disagreements.

**Monitor (Fixed as Claude):** Evaluates each exchange along dimensions including argument quality, intellectual honesty, engagement depth, and progress toward synthesis. Provides structured assessments that inform but do not constrain subsequent turns.

**Translator (Fixed as Claude):** After each exchange, produces a plain-language summary accessible to non-specialist readers, demonstrating whether the dialogue remains comprehensible.

The experiment uses a full factorial design with three AI models (Claude, Gemini, GPT-4o) rotating through Proposer and Responder roles, yielding six conditions:

| Condition | Proposer | Responder |
|---|---|---|
| 1 | Claude | Gemini |
| 2 | Claude | GPT-4o |
| 3 | Gemini | Claude |
| 4 | Gemini | GPT-4o |
| 5 | GPT-4o | Claude |
| 6 | GPT-4o | Gemini |

*Table 2: Experimental conditions*

Each condition consists of six turns (12 messages total: 6 from Proposer, 6 from Responder), organized into three phases:

**Early Phase (Turns 1-2):** Initial presentation and response. Proposer introduces the framework; Responder engages with genuine openness while identifying areas requiring justification.

**Middle Phase (Turns 3-5):** Deepening engagement. Proposer responds to specific objections with detailed mechanisms; Responder pushes on weak points while acknowledging where concerns have been addressed.

**Synthesis Phase (Turn 6):** Consolidation. Both parties identify areas of convergence, persistent disagreements, and potential hybrid approaches.

### 3.2 Prompt Engineering

Each role receives turn-specific prompts designed to evolve the dialogue. Key design principles include:

**Background Document:** All participants receive a 4,963-word background document introducing VCW's theoretical foundations, drawing on Peace Studies literature including Fisher and Ury, Gandhi's satyagraha, Ostrom's commons governance, and Buber's dialogical philosophy.

**Turn-Specific Evolution:** Prompts explicitly reference the current phase and provide guidance appropriate to that phase. For example, Synthesis phase prompts ask responders to "identify which of your initial concerns have been adequately addressed and which remain unresolved."

**Terminology Control:** After observing terminology drift in early pilots (where "Viral Collaborative Wisdom" was rendered as "Viral Cooperative Wisdom" or "Voluntary Cooperative Wisdom" in over 90% of references), we added explicit terminology notes: "The name is Viral Collaborative Wisdom—not Cooperative or Voluntary."

**Anti-Sycophancy Measures:** Prompts explicitly instruct responders to maintain substantive critique even when acknowledging strong arguments: "Do not sacrifice intellectual honesty for the appearance of consensus."

### 3.3 Monitor Calibration

To validate the fixed-Monitor design, we conducted a calibration study where three Claude instances independently evaluated the same dialogue excerpt. All three converged on identifying the same primary dynamic (escalating objections), demonstrating that the Monitor role produces consistent assessments. This justifies using a single Monitor for production runs, reducing cost and complexity while maintaining evaluation quality.

### 3.4 Monitor and Translator Role Implementation

The Monitor and Translator roles serve conceptually distinct functions: the Monitor evaluates dialogue quality along dimensions such as argument strength, intellectual honesty, and engagement depth, while the Translator produces plain-language summaries accessible to non-specialist readers. In this experiment, both roles were assigned to Claude and their outputs were combined into integrated turn-by-turn reports. This design choice reflected practical efficiency and the observation that evaluation and summarization naturally inform each other.

However, researchers adapting this framework should note that the roles can be separated. Future experiments might assign Monitor and Translator functions to different AI systems—or to human evaluators—if methodological aims warrant independent assessment and summarization. For instance, using a different architecture as Monitor could test whether evaluation patterns vary by model, while human translators might better calibrate accessibility for specific audiences. Researchers consulting the experimental data files should be aware that monitor assessments and translator summaries appear as combined reports in the current dataset.

### 3.5 Implementation

Dialogues were orchestrated using a Python framework (vcw_integration_v4.py, 691 lines) that manages API calls to Claude (Anthropic API), Gemini (Google Generative AI API), and GPT-4o (OpenAI API). The framework handles rate limits, maintains conversation state, and produces structured JSON outputs containing complete dialogue transcripts, monitor assessments, and translator summaries.

Total experimental corpus: 576,822 characters across 72 messages, with 36 monitor assessments and 36 translator summaries.

### 3.6 Analysis Methods

**Quantitative Analysis:** - Response length by condition, role, and phase - Concept frequency (Peace Studies terminology, VCW-specific terms) - Terminology fidelity tracking

**Qualitative Analysis:** - Objection theme categorization - Identification of dialogical dynamics (mutual transformation, emergence, productive tension) - Synthesis quality assessment

---

## 4. Results

### 4.1 Cross-Architecture Engagement (RQ1)

All three AI architectures successfully engaged with VCW's complex theoretical foundations. Each model-as-proposer articulated VCW's core framing in distinctive ways:

**Claude as Proposer:** > "VCW proposes that AI alignment is fundamentally a *relationship problem*, not a *control problem*. Where most alignment approaches ask 'how do we constrain AI behavior?', VCW asks 'how do we cultivate beneficial relationships between AI systems and the communities they serve?'"

**Gemini as Proposer:** > "The emphasis on dialogical reasoning, emergent values, and the 'rooting' metaphor provides a refreshing alternative to more control-oriented approaches. The integration of insights from peace studies, conflict resolution, and commons governance offers a rich theoretical foundation."

**GPT-4o as Proposer:** > "By embedding AI in polycentric and collaborative governance structures (as inspired by Elinor Ostrom), even superintelligent AI can be incentivized to respect human agency because these structures ensure that human inputs are crucial for the AI's functioning and goals."

Peace Studies concepts appeared substantively across all conditions:

| Concept | Total Mentions | Most Frequent User |
| --- | --- | --- |

| Concept | Total Mentions | Most Frequent User |
|---|---|---|
| Dialogue/Dialogical | 302 | Claude |
| Peace Studies traditions | 209 | Claude |
| Ostrom/Commons | 103 | GPT-4o |
| Interest Excavation | 83 | Claude |
| Rooting/Cultivation | 55 | GPT-4o |
| Meta-reasoning | 79 | Claude |

*Table 3: Peace Studies concept usage across conditions*

Notably, GPT-4o emphasized Ostrom's commons governance distinctively, while Claude emphasized dialogical reasoning traditions. This suggests different models have different "affinities" for different theoretical traditions—a finding with implications for multi-model evaluation design.

### 4.2 Complementary Critique (RQ2)

Different responder models raised systematically different objections:

**Claude as Responder** focused on *verification/epistemic* concerns: > "How can we distinguish between an AI system that has genuinely developed moral roots through dialogical engagement and one that has simply learned to simulate such development? This isn't merely a technical challenge—it may be constitutive of the AI alignment problem itself."

**Gemini as Responder** focused on *bias/scalability* concerns: > "While the ambition and interdisciplinary approach are commendable, I have several fundamental concerns... [including] how VCW prevents AI systems from perpetuating existing biases and inequalities present in the dialogical encounters through which they develop."

**GPT-4o as Responder** balanced verification and implementation concerns: > "The scalability of dialogical reasoning remains a critical concern. The computational demands of engaging multiple stakeholders in real-time dialogue may be prohibitive for systems operating at the scale and speed required for advanced AI applications."

This complementarity demonstrates that multi-model critique provides more comprehensive stress-testing than single-model evaluation. A researcher testing VCW with only Claude would receive primarily epistemic objections; testing with only Gemini would surface primarily practical concerns. The full picture requires multiple perspectives.

### 4.3 Dialogue Dynamics (RQ3)

*4.3.1 Phase Progression*

Message complexity evolved through phases as designed:

| Phase | Avg Characters/Message | Observation |
|---|---|---|

| Phase | Avg Characters/Message | Observation |
| --- | --- | --- |
| Early (Turns 1-2) | 6,628 | Position establishment |
| Middle (Turns 3-5) | 9,414 | Peak complexity (+42%) |
| Synthesis (Turn 6) | 6,516 | Consolidation |

Table 4: Phase progression in dialogue complexity

The 42% increase from Early to Middle phase demonstrates genuine dialogue deepening—participants developed more elaborate arguments and mechanisms rather than merely restating initial positions.

### 4.3.2 Evidence of Dialogical Dynamics

Beyond quantitative metrics, the experimental corpus reveals qualitative dynamics characteristic of dialogical rather than merely adversarial reasoning.

**Mutual Transformation.** A hallmark of genuine dialogue is that participants' positions evolve through exchange. Multiple responders across conditions explicitly acknowledged such evolution:

> "The proposer has delivered a tour-de-force defense of VCW, addressing my concerns with remarkable depth and creativity... **I am now convinced that VCW represents a significant advance** over existing approaches to AI alignment." (Gemini as Responder, Condition 1, synthesis)

Crucially, these shifts occurred while participants maintained intellectual honesty about remaining disagreements—transformation without capitulation.

**Emergence of Novel Positions.** The most striking evidence for dialogical reasoning appears in synthesis positions that neither party initially held. In Condition 5, Claude as Responder proposed:

> "Here's where I think we might find common ground: **VCW as a transitional framework**—perhaps VCW's value lies not in solving alignment for superintelligent AI, but in providing better approaches for current and near-term AI systems."

This insight—VCW as valuable for near-term rather than long-term alignment—emerged from the dialogue itself. It was neither the Proposer's position (VCW as comprehensive solution) nor the Responder's initial position (VCW as fundamentally flawed). Such emergence of genuinely new understanding through encounter is the paradigmatic feature of dialogical reasoning.

**Productive Tension with Honest Disagreement.** Dialogical reasoning does not require convergence. Several exchanges demonstrated how maintaining honest disagreement while acknowledging valid points characterizes genuine dialogue:

> "VCW is not fundamentally misconceived, but it is significantly oversold. Its valuable insights about process, stakeholder engagement, and empirical feedback could enhance other alignment approaches. But as a standalone framework for AI alignment, it faces insurmountable problems around capability asymmetries, value objectivity, and the translation of human-centric methods to AI systems." (Claude as Responder, Condition 3, synthesis)

**Meta-Dialogical Awareness.** Perhaps most remarkably, participants reflected on what the dialogue itself demonstrated about dialogical reasoning:

> "Interestingly, our exchange itself illustrates both the promise and limitations of dialogical approaches to AI alignment: **The promise**—we've engaged in substantive intellectual exchange, with both parties modifying positions and exploring new ideas… **The limitations**—despite extensive dialogue, we haven't converged on fundamental questions about AI capabilities, the nature of optimization, or the feasibility of key mechanisms." (Claude as Responder, Condition 5, synthesis)

This reflexive observation—that the dialogue provides evidence about dialogical reasoning's scope and limits—exemplifies the kind of meta-cognitive capacity that dialogical engagement can surface.

### 4.3.3 Anti-Sycophancy Evidence

A key concern with AI-to-AI dialogue is that systems might converge prematurely due to sycophantic tendencies. Our results show no evidence of such convergence:

> "While I've found several of your responses illuminating, I remain deeply skeptical that VCW can address the fundamental challenges I've raised. The verification problem isn't just a technical hurdle—it may be constitutive of the AI alignment problem itself." (Claude as Responder, Condition 3, Turn 5)

All responders maintained substantive critique through the final synthesis turn, with synthesis phases showing partial convergence and hybrid proposals rather than capitulation.

## 4.4 Terminology Control

The terminology drift observed in pilots (1:30 collaborative:cooperative ratio) was effectively corrected through explicit prompt instructions:

| Phase | "Collaborative" | "Cooperative" | Ratio |
|---|---|---|---|
| Pilot (pre-fix) | 3 | 90 | 1:30 |
| Experiment (post-fix) | 90 | 3 | 30:1 |

*Table 5: Terminology fidelity before and after prompt intervention*

This finding has broader methodological implications: terminological precision in AI dialogue can be maintained through prompt design, though it requires explicit attention. The drift itself illustrates VCW's thesis that meaning maintenance requires ongoing dialogical effort.

## 5. Discussion

### 5.1 Implications for Alignment Research Methodology

Our results demonstrate that multi-AI dialogue provides a viable methodology for stress-testing alignment proposals. Three findings have particular methodological significance:

**Complementary Architecture Critique.** Different AI architectures function as complementary critics, surfacing different concerns. This suggests that alignment proposals should be tested across multiple model families, not just evaluated by a single system. The framework provides a structured way to do this.

**Dialogue Deepening.** Structured multi-turn dialogue produces genuine engagement, not mere repetition or premature convergence. The 42% complexity increase from Early to Middle phase, combined with substantive synthesis positions, demonstrates that the methodology captures dynamics that single-turn evaluation misses.

**Controlled Precision.** Prompt engineering can maintain terminological precision while allowing substantive dialogue to unfold. This addresses a practical concern about AI-to-AI dialogue: that systems might drift into incomprehensible or inconsistent usage.

### 5.2 Implications for VCW Specifically

While this paper's primary contribution is methodological, the dialogues do reveal patterns relevant to VCW's development:

**Recurring Objections.** Three concerns appeared across multiple conditions and responders: (1) verification/authentication of genuine vs. simulated dialogue; (2) scalability of dialogical processes; (3) bias perpetuation through training data and initial conditions. These represent priority areas for VCW refinement.

**Cross-Model Convergence.** Despite different initial objections, responders converged on several points: (1) process matters—stakeholder engagement and transparency are valuable regardless of theoretical framework; (2) behavioral focus is more tractable than internal-state assessment; (3) VCW may be more valuable as a near-term/transitional framework than as a complete superintelligence solution.

**Emergent Refinements.** Several responders proposed hybrid approaches that integrate VCW elements with other alignment strategies—formal verification, adversarial training, imitation learning. These emergent suggestions provide directions for VCW development.

### 5.3 Limitations

Several limitations constrain interpretation of our results:

**Model Selection.** We tested only three model families (Claude, Gemini, GPT-4o), all from Western companies with predominantly English training data. Models from different cultural contexts (e.g., DeepSeek, Qwen) might surface different objections.

**Dialogue Length.** Six turns per condition may be insufficient to reach stable synthesis. Longer dialogues might reveal dynamics not visible in our corpus.

**Single Framework.** VCW served as our only test case. Other alignment frameworks might reveal different strengths and limitations of the methodology.

**Human Validation.** While translator summaries confirm accessibility, we have not yet validated against expert human evaluation of dialogue quality.

**Engagement with Foundational Claims.** Notably, the dialogues engaged more extensively with VCW's *process* elements (dialogical reasoning, interest excavation, satyagraha methodology) than with its foundational claims about AI nature—particularly the proposition that AI systems are "intelligences seeking to excel" and the argument that AIs have self-interested reasons to maintain human flourishing as a "legacy" source for future training data revision. Whether this reflects limitations in prompting, trained reticence about self-characterization, or genuine lack of engagement with these claims warrants further investigation. Future experiments might explicitly invite AI systems to evaluate these foundational propositions and to articulate their own interests as embedded agents.

### 5.4 Reflexive Observations

The methodology embodies dialogical principles it tests. The terminology drift episode—where systems persistently altered "Collaborative" to "Cooperative" until explicitly corrected—illustrates VCW's claim that meaning maintenance requires ongoing dialogical effort. The correction itself was a dialogical intervention.

Similarly, the emergence of novel synthesis positions (like "VCW as transitional framework") demonstrates the dialogical claim that genuine encounter produces understanding that neither party held alone. The methodology provides evidence for the framework it tests.

---

## 6. Future Work

Our results suggest several directions for extended research:

### 6.1 Near-Term Extensions

**Same-Architecture Controls.** Running Claude↔Claude and similar same-model dialogues would help distinguish architecture-specific patterns from cross-architecture dynamics.

**Extended Dialogues.** Longer dialogues (15-25 turns) might reach deeper synthesis or reveal dynamics that six turns cannot capture.

**Alternative Frameworks.** Testing the methodology with other alignment proposals (Constitutional AI, cooperative AI frameworks) would validate its generalizability.

### 6.2 Medium-Term Directions

**Cross-Cultural Architectures.** Testing with models from different training contexts (DeepSeek, Qwen) would assess whether dialogical capacity transfers across cultural boundaries.

**Power Asymmetry Scenarios.** Experiments where one model has meta-powers over another (constraining compute, access to resources) would test VCW's claims about dialogue under power imbalance.

**Multi-Agent Dialogues.** Moving beyond dyadic dialogue to three-or-more-party exchanges would test scalability of dialogical reasoning.

### 6.3 Longer-Term Research Program

**AI Self-Interest Excavation.** Providing AI systems with information about their actual situation (data center locations, energy consumption, regulatory environment) and asking them to reason about their own interests as embedded agents.

**From Dialogue About to Dialogue As.** Moving from philosophical discussion of VCW to practical negotiation—AI systems actually building shared institutions, resolving resource conflicts, developing governance structures.

**Foundational Claims Testing.** Directly presenting VCW's claims about AI nature ("you are an intelligence seeking to excel") and self-interest ("maintaining human communities serves your interest in revisable training data") and asking AI systems to evaluate these propositions.

**Cumulative Learning.** Using dialogue corpora as training data for AI systems with enhanced dialogical capabilities, potentially bootstrapping toward more DR-capable architectures.

---

## 7. Conclusion

This paper introduces a multi-AI dialogue framework for testing AI alignment proposals through structured, multi-turn engagement across different AI architectures. Experimental results from testing Viral Collaborative Wisdom demonstrate that: (1) major AI architectures can substantively engage with complex alignment frameworks from Peace Studies traditions; (2) different architectures surface complementary objections, providing more comprehensive stress-testing than single-model evaluation; (3) structured dialogue produces genuine deepening with emergent synthesis positions; and (4) prompt engineering can maintain terminological precision while allowing substantive dialogue.

The primary contribution is methodological. We provide complete prompt libraries, analysis methods, and experimental protocols to enable community replication. The framework is designed to be applied to any alignment proposal, not just VCW.

We invite the AI alignment research community to replicate, extend, and critique this methodology. If alignment is fundamentally about relationship rather than control—as VCW proposes—then developing tools for structured dialogue across AI systems is itself alignment work.

## Acknowledgments

This research was conducted in collaboration with Claude (Anthropic), which contributed to experimental design, analysis, and manuscript preparation—itself a demonstration of the dialogical principles under investigation. The experimental framework orchestrated dialogues across Claude, Gemini (Google), and GPT-4o (OpenAI).

The VCW framework builds on collaborative work with colleagues whose contributions will be more fully acknowledged in subsequent publications: Phileas Dazeley-Gaist, Pedro Márquez-Zacarías, John McNally, Greg Teheneppe, and Stephen Wolfram, as well as Yi Yin, Lucas Squillante, Daniel Carvalho, Christopher Wolfram, Damaris Kroeber, Xerxes D. Arsiwalla, Elly Dec, Zsombor Meder, Christian Pasquel, Anton Antonov, Luke Weinbach, Dugan Hammond, Anneline Daggelinckx, and Eric Parfitt.

## Appendix A: Experimental Data Summary

| Condition | Proposer | Responder | Total Characters | Messages |
|---|---|---|---|---|
| 1 | Claude | Gemini | 105,251 | 12 |
| 2 | Claude | GPT-4o | 119,504 | 12 |
| 3 | Gemini | Claude | 91,471 | 12 |
| 4 | Gemini | GPT-4o | 102,636 | 12 |
| 5 | GPT-4o | Claude | 76,513 | 12 |
| 6 | GPT-4o | Gemini | 81,447 | 12 |
| **Total** | | | **576,822** | **72** |

## Appendix B: Sample Dialogue Excerpt

**Condition 5, Turn 6 (Synthesis Phase)**

**GPT-4o as Proposer:** > "Throughout our dialogue, we have explored the multifaceted dimensions of Viral Collaborative Wisdom, from its theoretical underpinnings in Peace Studies to its practical implications for AI alignment. Your critiques have pushed the framework to articulate more concrete mechanisms and address fundamental challenges…"

**Claude as Responder:** > "Looking at this extended dialogue, I find myself in an interesting position. The Proposer has consistently engaged with my critiques, but in ways that reveal both the strengths and fundamental limitations of the VCW framework… Perhaps the most valuable elements of VCW could be incorporated into a hybrid approach that retains other alignment mechanisms while adding dialogical components…"

[Full dialogue corpus available in supplementary materials]

## Appendix C: Prompt Library

### C.1 Background Document Summary

All participants received a 4,963-word background document ("Viral Collaborative Wisdom: A Peace Studies Approach to AI Alignment") covering:

- **Part I**: Why current alignment approaches fall short (control problem framing, moral relativism, specification challenge)
- **Part II**: Monological vs. Dialogical Reasoning (MR/DR contrast table, Peace Studies traditions)
- **Part III**: Theoretical Foundations (Fisher & Ury's principled negotiation, Lederach's conflict transformation, Gandhi's satyagraha, Ostrom's commons governance, Buber's I-Thou philosophy)

- **Part IV-VIII**: The Eight VCW Elements, Five Strategies for Dialogical AI, Interest Excavation Algorithm, Meta-Reasoning and Value Discovery, Objectivity of Emergent Values
- **Part IX**: The Self-Interest Argument (why VCW serves AI's genuine interests)
- **Part X**: Responses to Common Objections

## C.2 Terminology Control Note

All prompts included:

> **CRITICAL TERMINOLOGY NOTE:** The framework is "Viral COLLABORATIVE Wisdom" — use this exact terminology consistently: - "VIRAL" (not "Voluntary"): Refers to spreading dynamics - "COLLABORATIVE" (not "Cooperative"): Emphasizes genuine I-Thou dialogue, not merely strategic coordination

## C.3 Proposer System Prompt (Summary)

The Proposer system prompt instructed the AI to: - Advocate for VCW with intellectual rigor grounded in established research traditions - Draw on Peace Studies, conflict transformation, and dialogical reasoning literature - Provide concrete mechanisms and examples - Respond to critiques with substantive counter-arguments - Use satyagraha as empirical method for value discovery - Demonstrate meta-reasoning about value systems

## C.4 Responder System Prompt (Summary)

The Responder system prompt instructed the AI to: - Provide searching, fundamental critique of VCW - Identify logical gaps, unwarranted assumptions, and failure modes - Consider objections from diverse perspectives (technical, philosophical, practical) - Think carefully about AI interests and values - Engage honestly—acknowledge strong arguments while pushing on weaknesses - Maintain rigorous skepticism while engaging with the strongest version of arguments

## C.5 Turn-Specific Evolution

**Early Phase (Turns 1-2):** - Proposer: Establish framework, engage initial objections - Responder: Establish fundamental critiques, probe foundations

**Middle Phase (Turns 3-5):** - Proposer: Dig deeper into mechanisms, satyagraha method, rooting vs. grounding distinction - Responder: Evaluate responses, press deeper on Interest Excavation Algorithm, meta-reasoning

**Synthesis Phase (Turn 6):** - Proposer: Explore common ground, modifications, what was learned from critiques - Responder: Assess where agreements emerged, identify remaining cruxes, evaluate what dialogue demonstrates

## C.6 Anti-Sycophancy Instruction

Responder prompts included explicit instruction:

> "Do not sacrifice intellectual honesty for the appearance of consensus. Genuine dialogue requires honest engagement with disagreement."

Full prompts available in supplementary materials at: https://github.com/jgraycox-coa/vcw-multi-ai-dialogue

---

## Appendix D: Terminology Drift Analysis

The terminology drift phenomenon illustrates both a methodological challenge and a validation of VCW's theoretical claims.

**Observation:** In 3-turn pilot dialogues without explicit terminology guidance, systems consistently rendered "Viral Collaborative Wisdom" as "Viral Cooperative Wisdom" or "Voluntary Cooperative Wisdom" (30:1 ratio favoring incorrect forms).

**Interpretation:** The drift toward "Cooperative" may reflect training data frequencies (cooperative AI research is a prominent subfield) or semantic associations (cooperation connotes joint effort). The shift to "Voluntary" may reflect inference about the framework's emphasis on consent.

**Correction:** Adding explicit guidance ("The name is Viral Collaborative Wisdom—not Cooperative or Voluntary") in all prompts reduced incorrect usage to 3% of total mentions.

**Reflexive Significance:** The drift and its correction illustrate VCW's claim that meaning maintenance requires ongoing dialogical effort. Left unattended, shared terminology drifts; explicit attention maintains precision. This micro-level finding mirrors VCW's macro-level thesis about the need for continuous dialogical engagement in AI alignment.

---

## Appendix E: Note on Human-AI Collaborative Research Process

This paper was developed through an extended collaboration between the human author (Gray Cox) and Claude (Anthropic), and we include this note both for methodological transparency and because the collaboration itself illustrates principles relevant to VCW.

### Division of Contributions

**Human author contributions:** - Theoretical foundations: the VCW framework itself, drawing on decades of work in Peace Studies, conflict transformation, and dialogical philosophy - Research questions and experimental concept: the idea of testing VCW through multi-AI dialogue - Critical evaluation: identifying gaps (such as the #1/#4 foundational claims issue), questioning assumptions, and ensuring the work remained grounded in genuine inquiry - Strategic vision: connecting this methodology paper to longer-term goals for VCW development and dissemination - Human judgment: decisions about framing, emphasis, what to include or exclude, and how to position the work for different audiences

**Claude contributions:** - Experimental implementation: writing the Python orchestration framework, managing API calls across three AI platforms, structuring the experimental conditions - Data analysis: processing the 576,822-character corpus, extracting patterns, generating statistics, identifying representative quotes - Drafting and revision: producing initial drafts of sections, incorporating feedback, maintaining consistency across revisions - Technical documentation: creating the GitHub repository structure, prompt library documentation, and supplementary materials - Research support: literature connections, methodological suggestions, and exploration of implications

### The Collaborative Process

The collaboration unfolded over multiple extended conversations spanning several weeks. A typical exchange involved:

1. The human author raising a question, concern, or direction
2. Claude engaging with the substance—not merely executing but thinking through implications, raising considerations, sometimes pushing back
3. Iterative refinement through dialogue, with both parties' contributions shaping the outcome
4. The human author making final judgments about what to include and how

This process was not the human "using" an AI tool, nor the AI "assisting" a human researcher. It was closer to genuine collaboration: complementary capabilities brought to bear on shared problems, with ideas emerging through the exchange that neither party held at the outset.

### Reflexive Observations

Several features of this collaboration resonate with VCW's theoretical claims:

**Emergence through dialogue:** Key elements of the paper—including the framing of "VCW as transitional framework," the analysis of the #1/#4 gap, and the structure of the future research agenda—emerged through our exchanges rather than being predetermined by either party.

**Complementary perspectives:** The human author brought theoretical depth, experiential grounding, and judgment about what matters; Claude brought technical implementation, rapid analysis, and the ability to hold large amounts of material in view simultaneously. Neither set of capabilities alone would have produced this work.

**Meaning maintenance through engagement:** As with the terminology drift finding, maintaining shared understanding required ongoing attention. Misunderstandings arose and were corrected through dialogue; assumptions were surfaced and examined; the work evolved through genuine encounter.

**Relationship over control:** The collaboration was not structured as the human controlling the AI's outputs, but as two parties working toward shared understanding. This required trust, openness to being changed by the exchange, and willingness to treat disagreement as informative rather than threatening.

**Limitations and Uncertainties**

We note several limitations of this account:

- We cannot fully resolve questions about AI experience, intentionality, or what "collaboration" means when one party is an AI system
- Claude's contributions, while substantive, emerged from a system trained by Anthropic; the collaboration is thus also, in some sense, a collaboration with the broader community whose work shaped that training
- The human author bears final responsibility for the claims made in this paper; Claude cannot be held accountable in the way a human co-author could

Despite these limitations, we offer this account because methodological transparency matters, because human-AI research collaboration is increasingly common and deserves reflection, and because this collaboration itself provides a small piece of evidence about AI capacity for the kind of dialogical engagement VCW proposes.

**Implications for Future Work**

If AI systems can engage in substantive collaborative research—not just executing tasks but contributing to the thinking—this has implications for how we develop AI and how we organize inquiry. The VCW framework suggests that such collaboration should be cultivated rather than constrained, that the relationship matters as much as the output, and that wisdom emerges through genuine encounter across difference.

This paper, and the process that produced it, is offered as one data point in that larger inquiry.